%% file: hypernomad_v7_generic.tex
\documentclass[12pt,english]{article}
\usepackage{hyperref}
\usepackage{breakurl}

\hypersetup{
    unicode = false,
    pdftoolbar = true,
    pdfmenubar = true,
    pdffitwindow = true,
    pdftitle = {HYPERNOMAD},
    pdfauthor = {Lakhmiri, Le Digabel, Tribes},
    pdfsubject = {HYPERNOMAD},
    pdfnewwindow = true,
    pdfkeywords = {HPO, MADS},
    colorlinks = true,
    linkcolor = blue,
    citecolor = blue,
    filecolor = black,
    urlcolor = blue,
    breaklinks = true
}

\DeclareFixedFont{\ttb}{T1}{txtt}{bx}{n}{12} % for bold
\DeclareFixedFont{\ttm}{T1}{txtt}{m}{n}{12}  % for normal

\usepackage{color}
\definecolor{deepblue}{rgb}{0,0,0.5}
\definecolor{deepred}{rgb}{0.6,0,0}
\definecolor{deepgreen}{rgb}{0,0.5,0}

\usepackage{makecell}
\usepackage[english]{babel}
\usepackage[latin1,applemac]{inputenc}
\usepackage{amsmath}
\usepackage{amssymb}
\usepackage{multirow}
\usepackage{soul}
\usepackage{colortbl}
\usepackage{geometry}
\usepackage{times}
\usepackage{xspace}
\usepackage{graphicx}
\usepackage{algorithm}
\usepackage{wrapfig}
\usepackage{listings}
\usepackage{longtable}
\usepackage{subcaption}

\usepackage{appendix}

\usepackage{tikz}
\usetikzlibrary{fit, arrows, calc, positioning}
\usetikzlibrary{arrows, automata}
\usetikzlibrary{positioning}
\usepackage{array}

\tikzset{
    state/.style={
           rectangle,
           rounded corners,
           draw=black, very thick,
           minimum height=2em,
           inner sep=2pt,
           text centered,
           },
}

\tikzstyle{b} = [rectangle, draw, fill=blue!20, node distance=3cm, text width=6em, text centered, rounded corners, minimum height=4em, thick]
\tikzstyle{c} = [rectangle, draw, minimum height=20em, minimum width=10em, dashed]
\tikzstyle{l} = [draw, -latex',thick]

\usepackage{booktabs}

\usepackage{pifont}

\newcommand{\cmark}{\ding{51}}%

\def\R{{\mathbb{R}}}
\def\Z{{\mathbb{Z}}}
\def\deltA{{\boldsymbol \delta}}
\def\DeltA{{\boldsymbol \Delta}}
\newcommand{\diag}{\mathop{\mathrm{diag}}}

\definecolor{Red}{rgb}{1,0,0}
\definecolor{Gray}{rgb}{0.2,0.2,0.2}
\definecolor{Maroon}{rgb}{0.6,0.05,0.03}
\definecolor{Blue}{rgb}{0,0.7,0.9}
\definecolor{Green}{rgb}{0,.7,0}

\newcommand{\vizier}{{\sf Google Vizier}\xspace}
\newcommand{\mads}{{MADS}\xspace}
\newcommand{\nomad}{{\sf NOMAD}\xspace}
\newcommand{\hypernomad}{{\sf HyperNOMAD}\xspace}
\newcommand{\hyperopt}{{\sf hyperopt}\xspace}
\newcommand{\pytorch}{{\sf PyTorch}\xspace}
\newcommand{\hypernomadurl}{https://github.com/DouniaLakhmiri/HyperNOMAD}

\newcommand\pythoninline[1]{{\pythonstyle\lstinline!#1!}}

\begin{document}

%------------------------------------------------%
\title{
    \hypernomad: Hyperparameter optimization of deep neural networks using mesh adaptive direct search\thanks{
    {GERAD}
          and D\'epartement de Math\'ematiques et G\'enie Industriel,
          Polytechnique Montr\'eal,
          C.P. 6079, Succ. Centre-ville,
          Montreal, QC, Canada H3C 3A7.
   }
}

\author{
	   \href{mailto:Dounia.Lakhmiri@gerad.ca}{Dounia Lakhmiri}\thanks{
	             \href{Dounia.Lakhmiri@gerad.ca}{Dounia.Lakhmiri@gerad.ca}.
	}
\and	     
       \href{mailto:Sebastien.Le.Digabel@gerad.ca}{S\'ebastien Le~Digabel}\thanks{
          \href{https://www.gerad.ca/Sebastien.Le.Digabel}{www.gerad.ca/Sebastien.Le.Digabel}.
   	}
\and
    \href{mailto:Christophe.Tribes@gerad.ca}{Christophe Tribes}\thanks{
    	\href{mailto:Christophe.Tribes@gerad.ca}{Christophe.Tribes@gerad.ca}.
  }
}

\maketitle

\vspace*{-0.5cm}

\noindent
{\bf Abstract.}
The performance of deep neural networks is highly sensitive to the choice of the hyperparameters that define the structure of the network and the learning process. When facing a new application, tuning a deep neural network is a tedious and time consuming process that is often described as a ``dark art''. This explains the  necessity of automating the calibration of these hyperparameters. Derivative-free optimization is a field that develops methods designed to optimize time consuming functions without relying on derivatives.
This work introduces the \hypernomad package, an extension of the \nomad software that applies the \mads algorithm~\cite{AuDe2006} to simultaneously tune the hyperparameters responsible for both the architecture and the learning process of a deep neural network (DNN), and that allows for an important flexibility in the exploration of the search space by taking advantage of categorical variables.
This new approach is tested on the MNIST and CIFAR-10 data sets and achieves results comparable to the current state of the art. \\



% http://www.ams.org/msc/msc2010.html
%
%	90C30  	Nonlinear programming
%	90C56  	Derivative-free methods and methods using generalized derivatives [See also 49J52]
%	65K05  	Mathematical programming methods [See also 90Cxx]
%	62P30  	Applications in engineering and industry

%------------------------------------------------%
%------------------------------------------------%
%------------------------------------------------%

\input{manuscript.tex}

%------------------------------------------------%
%------------------------------------------------%
%------------------------------------------------%

%------------------------------------------------%
% References
%------------------------------------------------%
\bibliographystyle{plain}
\bibliography{bibliography}
\pdfbookmark[1]{References}{sec-refs}
% \label{sec-refs}
%------------------------------------------------%

%------------------------------------------------%
% Appendix
%------------------------------------------------%
\appendixpage
\appendix
\section[Using HyperNOMAD]{Using \hypernomad}
\label{sec-appendix}

\hypernomad is a {\sf C++} and {\sf Python} package dedicated to the hyperparameter optimization of deep neural networks. The package contains a blackbox specifically designed for this problematic and provides a link with the \nomad software~\cite{Le09b} used for the optimization. The blackbox takes as inputs the hyperparameters discussed in Section~\ref{sec:hpos}, builds a corresponding deep neural network in order to train, validate and test it on a specific data set before returning the test accuracy as a mesure of performance. \nomad is then used to minimize this error. The following appendix provides an overview of how to use the \hypernomad package.

%------------------------------------------------%
\subsection*{Prerequisites}
%------------------------------------------------%

\hypernomad relies on: 
\begin{itemize}
	\item A compiled version of the \nomad software available at \url{https://www.gerad.ca/nomad/} for the optimization;
	\item The \pytorch library available at \url{https://pytorch.org/} for modeling the neural network within the blackbox;
	\item A version of {\sf Python} superior to~3.6;
	\item A version of {\sf gcc} superior to~3.8.
\end{itemize}

\subsection*{Installation of \hypernomad}

\hypernomad is available at
\href{https://github.com/DouniaLakhmiri/HyperNOMAD}{
{\tt https://github.com/DouniaLakhmiri/Hyper} {\tt NOMAD}
}. The user must produce the executable \texttt{hypernomad.exe} using the provided makefile as follows:

\begin{lstlisting}
> make
	building HYPERNOMAD ...

    	To be able to run the example
        the HYPERNOMAD_HOME environment variable
        must be set to the HyperNOMAD home directory
\end{lstlisting} 

\vspace*{5mm}
When the compilation is successful, a message appears asking to set an environment variable {\tt HYPERNOMAD\_HOME} which can be done by adding a line in the 
{\tt .profile} or {\tt .bashrc} files:

\begin{lstlisting}
export HYPERNOMAD_HOME=hypernomad_directory
\end{lstlisting}

\vspace*{5mm}
The user can check that the installation is successful by trying to run the command:\\

\begin{lstlisting}
> $HYPERNOMAD_HOME/bin/hypernomad.exe -i

--------------------------------------------------
  HYPERNOMAD - version 1.0
--------------------------------------------------
  Using Nomad version 3.9.0 - www.gerad.ca/nomad
--------------------------------------------------

Run           : hypernomad.exe parameters_file
Info          : hypernomad.exe -i
Help          : hypernomad.exe -h
Version       : hypernomad.exe -v
Usage         : hypernomad.exe -u
Neighboors    : hypernomad.exe -n parameters_file

\end{lstlisting}

%------------------------------------------------%
\subsection*{Using \hypernomad}
%------------------------------------------------%

The next phase is to create a parameter file that contains the necessary information to specify the classification problem, the search space and the initial starting point. \hypernomad allows for a good flexibility of tuning a convolutional network by considering multiple aspects of a network at once such as the architecture, the dropout rate, the choice of the optimizer and the hyperparameters related to the optimization aspect (learning rate, weight decay, momentum, etc.), the batch size, etc. The user can choose to optimize all these aspects or select a few and fix the others to certain values. The user can also change the default range of each hyperparameter. This information is passed through the parameter file by using a specific syntax where ``LB" represents the lower bound and ``UB" the upper bound.

\begin{lstlisting}
KEYWORD  INITIAL_VALUE  LB  UB  FIXED/VAR
\end{lstlisting}

\vspace*{5mm}While the hyperparameters have default values in \hypernomad, the data set must be explicitly provided by the user in a separate file in order to specify the considered optimization problem. The following section explains how to specify the necessary parameter file before running an optimization.

%------------------------------------------------%
\subsubsection*{Choosing a data set}
%------------------------------------------------%

The library can be used on different data sets whether they are already incorporated in \hypernomad, such as the ones listed in Table~\ref{tab:datasets}, or are provided by the user. In the latter case, please refer to the user guide in \url{https://hypernomad.readthedocs.io/en/latest/} for details on how to link a personal data set to the library. The rest of the section describes how to run an optimization on a data set provided with \hypernomad.

Because of the nature of the applications considered by \hypernomad, the computing time can become constraining, especially during the training phase of each configuration, which is why ``{\sf TOYMNIST}" is created as a subset of MNIST containing 300 training images, 100 for the validation and another 100 for testing. It is added to the package for experimenting with \hypernomad without having to wait several hours for each blackbox evaluation. 

%------------------------------------------------%
\subsubsection*{Specifying the search space}
%------------------------------------------------%

In order to specify the problem to optimize and its parameters, the user must provide a parameter file that contains all the necessary informations to run an optimization.  As shown below, the parameter file consists of a list of keywords, each corresponding to a hyperparameter, and the values that the user wishes to attribute them. Some of these key words are mandatory such as the data set, in order to specify the problem, and the number of blackbox evaluations. Other keywords are optional and have default values if they do not appear on the parameter file. Table~\ref{tab:keywords} summarizes all the possible keywords with their default values and ranges. The user can change the lower and upper bounds of a hyperparameter and decide to maintain a hyperparameter to a fixed value during the entire optimization.

Below is a first example of a parameter file that corresponds to the one provided
in {\tt \$HYPERNOMAD\_HOME/examples/mnist\_first\_example.txt}.
First, the MNIST data set is chosen and \hypernomad is allowed to try a maximum of 100 configurations. Then, the number of convolutional layers is fixed throughout the optimization to five. The two ``{\tt -}" appearing after the ``{\tt 5}" mean that the default lower and upper bounds are maintained. The kernels, number of fully connected layers, and activation function, are respectively initialized to 3, 6, and 2 (Sigmoid) and the dropout rate is initialized to 0.6 with a new lower bound of 0.3 and upper bound of 0.8  instead of the default range of [0;1]. Finally, all the remaining hyperparameters from Table~\ref{tab:keywords} that are not explicitly mentioned in this file are fixed to their default values.

\begin{lstlisting}
# Mandatory information
DATASET                 MNIST
MAX_BB_EVAL             100

# Optional information
NUM_CON_LAYERS          5  -  -  FIXED # The initial value is fixed
                                       # lower and upper bounds have
                                       # no influence when parameter 
                                       # is fixed.

KERNELS                 3 # Only the initial value is set (not fixed)
                          # the lower bound and upper bound
                          # have default values.

NUM_FC_LAYERS           6
ACTIVATION_FUNCTION     2
DROPOUT_RATE            0.6  0.3 0.8  # The lower and upper bounds 
                                      # are set to values that are not 
                                      # the default ones
REMAINING_HPS           FIXED
\end{lstlisting}

\vspace*{5mm}
Below is a second example of a parameter file where the user is only interested in optimizing the fully connected block of a CNN on a the MNIST data set. All the remaining aspects of the network are fixed to their default values throughout the execution of \hypernomad. The optimization starts from a point with 10 fully connected layers of the same size of 500 neurones. This parameter file is provided with the package in \texttt{\$HYPERNOMAD\_HOME/examples/mnist\_fc\_optim.txt}.

\begin{lstlisting}
# Mandatory information
DATASET                 MNIST
MAX_BB_EVAL             150

# Optional information
NUM_FC_LAYERS           10  # Initial value is set to 10
                            # the lower and upper bounds are 
                            # the default ones
                            
SIZE_FC_LAYER           500  -  2000 # Initial value is set to 500
                                     # the lower bound is the default one
                                     # the upper bound in now 2000
                                     
REMAINING_HPS           FIXED
\end{lstlisting}

\vspace*{5mm}Finally, below is a minimal parameter file where only the mandatory information is specified. The execution of \hypernomad starts from the default starting point and all the hyperparameters of Table~\ref{tab:keywords} can be changed. The last line of this file can actually be removed without changing the behavior of \hypernomad since the default value for \texttt{REMAINING\_HPS} is set to \texttt{VAR}. Executing \hypernomad with this file should return the same values obtained in Figure~\ref{fig:comp}. This file is provided in \texttt{\$HYPERNOMAD\_HOME/examples/cifar10\_default.txt}.

\begin{lstlisting}
# Mandatory information
DATASET                 CIFAR10
MAX_BB_EVAL             100

REMAINING_HPS           VAR
\end{lstlisting}

%------------------------------------------------%
\subsubsection*{Running an execution}
%------------------------------------------------%

\begin{figure}[ht]
   \centering
       \includegraphics[page=1,width=.5\textwidth]{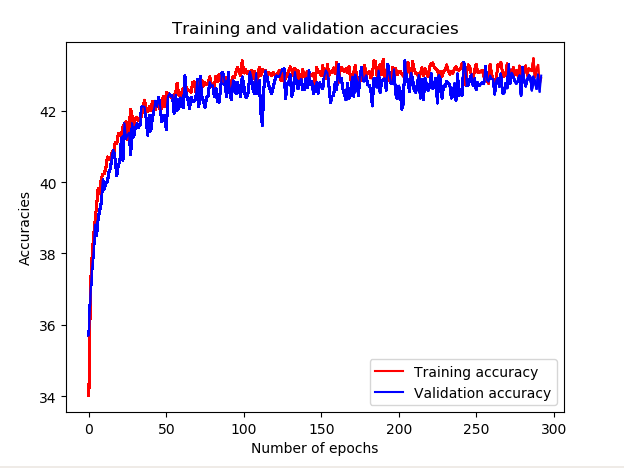} \\ 
 \caption{Example of a window that appears during one evaluation of the blackbox in \hypernomad. This figure shows in real time the training and validation accuracies of the current evaluated set of hyperparameters, at each epoch.}
 \label{fig:training}
\end{figure}

The user can run the previous example  by executing the following command from the {\tt examples} directory:
\begin{lstlisting}[breaklines=true]
> $HYPERNOMAD_HOME/bin/hypernomad.exe ./mnist_fc_optim.txt
\end{lstlisting}

During the optimization, a window appears to plot the training and validation accuracies of the network corresponding to the current point at each epoch as shown in Figure~\ref{fig:training}. When the optimization is done, \hypernomad produces the two files \texttt{history.txt} and \texttt{stats.txt}. The first contains each evaluated point and the corresponding testing accuracy, and the second contains the list of successful points. 

\begin{table}[ht]
\caption{Keywords for the \hypernomad parameters file.}
  \centering
  \resizebox{0.89\textwidth}{!}{
   \begin{tabular}{lccr} \toprule
   {Name } & {Description} & \makecell{Default \\ value} & \makecell{Scope} \\ \midrule
   {\tt DATASET} 	   	&   \makecell{Name of the data set \\used for the optimization}	     & No default	 & \makecell{ A data set from Table~\ref{tab:datasets} or\\ {\tt CUSTOM} for a custom data set}\\ \midrule
   {\tt NUMBER\_OF\_CLASSES}&   \makecell{Number of classes of \\ the classification problem}	     & \makecell{No default. Only use if \\ {\tt DATASET} $=$ {\tt CUSTOM}}	          & $\mathbb{N}$ \\ \midrule
   {\tt MAX\_BB\_EVAL}  	   	&  \makecell{Maximum number of calls\\ to the blackbox}	     &  No default  & [1; $\infty$]\\ \midrule
   {\tt NUM\_CON\_LAYERS} 	   		&  Number of convolutional layers 	     & {\tt 2}	          & [0;100]\\ \midrule
   {\tt OUTPUT\_CHANNELS}	& \makecell{Number of output channels\\ for each convolutional layer}  & {\tt 6} & [1;100] \\ \midrule
   {\tt KERNELS}	   		&  \makecell{Size of the kernel applied \\to each convolutional layer}	     & {\tt 5}	          & [1;20]\\ \midrule
   {\tt STRIDES} 	   		&  \makecell{Step of the kernel for\\ each convolutional layer}     & {\tt 1}	          & [1;3]\\ \midrule
   {\tt PADDINGS} 	   		&  \makecell{Size of the padding for \\  each convolutional layer}     & {\tt 0}	          & [0;2]\\ \midrule
   {\tt DO\_POOLS}			&  \makecell{Apply a pooling after \\  each convolutional layer}	     & {\tt 0}	          & $\{$0,1$\}$\\  \midrule
   {\tt NUM\_FC\_LAYERS} 	   		&  Number of fully connected layers 	     & {\tt 2}	          & [0;500]\\ \midrule
   {\tt SIZE\_FC\_LAYER} 	& Size of each fully connected layer & {\tt 128} & [1;1,000]\\ \midrule
   {\tt BATCH\_SIZE} 	& \makecell{Size of batch for the \\mini-batch gradient}	 & {\tt 128} & [1;400] \\ \midrule
   {\tt OPTIMIZER\_CHOICE} & \makecell{Optimizer to use \\ from Table~\ref{tab:optimizer}} & {\tt 3} & $\{$1,2,3,4$\}$ \\ \midrule
   {\tt OPT\_PARAM\_1} & \makecell{Learning rate} & {\tt 0.1}  & [0;1]\\ \midrule
   {\tt OPT\_PARAM\_2} & \makecell{Second hyperparameter \\ related to the optimizer.} &  {\tt 0.9}  & [0;1]\\ \midrule
   {\tt OPT\_PARAM\_3} & \makecell{Third hyperparameter \\related to the optimizer.} &  {\tt 0.005} &  [0;1]\\ \midrule
   {\tt OPT\_PARAM\_4} & \makecell{Fourth hyperparameter \\related to the optimizer.} &  {\tt 0}  &  [0;1]\\ \midrule
   {\tt DROPOUT\_RATE} & \makecell{Probability that a node \\ will be dropped out} & {\tt 0.5} & [0;0.95] \\ \midrule
   {\tt ACTIVATION\_FUNCTION} &  \makecell{Choice of the activation function\\ from ReLU, Sigmoid and Tanh} & {\tt 1} & $\{$1,2,3$\}$ \\ \midrule
   {\tt REMAINING\_HPS} & \makecell{ Allows to fix or to vary all the \\  hyperparameters not explicitly \\ mentioned in the parameter file } & {\tt VAR} &
   $\{$ {\tt FIXED} , {\tt VAR}$\}$ \\
   \bottomrule
   \end{tabular}}
   \label{tab:keywords}
\end{table}

%------------------------------------------------%
\end{document}

%% file: manuscript.tex
%------------------------------------------------%
\section{Introduction}
%------------------------------------------------%

%%%%%%%%%%%
%      NN
%%%%%%%%%%%
Neural networks are mathematical structures used to solve supervised classification problems such as images, sounds and speech, to name a few. In the recent years, neural networks gained in popularity and were declined in different versions: deep, convolutional, recurrent, etc. in order to adapt to specific problematics. This popularity is due to the emergence of large size databases and the development of computational power of contemporary machines, through the use of GPUs in particular. These favorable conditions have allowed neural networks to learn complex structures and achieve a level of precision that can surpass human performance across multiple instances such as robotics~\cite{levine2018learning}, medical diagnostics~\cite{litjens2017survey}, and more.

However, the performance of a neural network is strongly linked to its structure and to the values of the parameters of
the optimization algorithm used to minimize the error between the predictions of the network and the data during its training.
The choices of the neural network hyperparameters can greatly affect its ability to learn from the training data and to generalize with new data. The algorithmic hyperparameters of the optimizer must be chosen a priori and cannot be modified during optimization. Hence, to obtain a neural network, it is necessary to fix several hyperparameters of various types: real, integer and categorical. A variable is categorical when it describes a class, or category, without a relation of order between these categories. The search for an optimal configuration is a very slow process that, along with the training, takes up the majority of the time when developing a network for a new application. It is a relatively new problem that is often solved randomly or empirically.

% To define a deep neural network, it is necessary to fix several hyperparameters of various types: real, integer and categorical. An algorithmic hyperparameter is a parameter that must be chosen a priori and cannot be modified during its execution. A variable is categorical when it describes a class, or category, without a relation of order between these categories. The search for an optimal configuration is a very slow process that, along with the training, takes up the majority of the time when developing a network for a new application. It is a relatively new problem that is often solved randomly or empirically. However, the choices of the neural network hyperparameters can greatly affect its ability to learn from the training data and to generalize with new data.

%%%%%%%%%%%
%	DFO 
%%%%%%%%%%%	
Derivative free optimization (DFO)~\cite{AuHa2017, CoScVibook} is the field that aims to solve optimization problems where the derivatives are unavailable, although they might exist. This is the case for example when the objective and/or constraints functions are non differentiable, noisy or expensive to evaluate. In addition, the evaluation in some points may fail especially if the values of the objective and/or contraints are the outputs of a simulation or an experience. Blackbox optimization (BBO) is a subfield of DFO where the derivatives do no exist and the problem is modeled as a blackbox. This term refers to the fact that the computing process behind the output values is unknown. The general DFO problem is described as follows:
\begin{align*}
	\min_{x  \in \Omega} f(x) 
\end{align*}
where $f$ is the objective function to minimize over the domain $\Omega$. 

There are two main classes of DFO methods: model-based and direct search methods.
The first uses the value of the objective and/or the constraints at some already evaluated points to build a model able to guide the optimization by relying on the predictions of the model. For example, this class includes methods based on trust regions~\cite[Chapter~10]{CoScVibook} or interpolations models~\cite{bobyqa}. This differentiates them from direct search methods~\cite{HoJe61a} that adopt a more straightforward strategy to optimize the blackbox. At each iteration, direct search methods generate a set of trial points that are compared to the ``best solution" available. For example, the GPS algorithm~\cite{Torc97a} defines a mesh on the search space and determines the next point to evaluate by choosing a search direction. DFO algorithms usually include a proof of convergence that ensures a good quality solution under certain hypotheses on the objective function. BBO algorithms extend beyond this scope by including heuristics such as evolutionary algorithms, sampling methods and so on.

%%%%%%%%%%%
%	Link between the two
%%%%%%%%%%%

In~\cite{AuDaOr2014, AuOr06a}, the authors explain how a hyperparameter optimization (HPO) problem can be seen as a blackbox one. Indeed, the HPO problem is equivalent to a blackbox that takes the hyperparameters of a given algorithm and returns some measure of performance defined in advance such as the time of resolution, the value of the best point found or the number of solved problems. In the case of neural networks, the blackbox can return the accuracy on the test data set as a mesure of performance. With this formulation, DFO techniques can be applied to solve the original HPO problem.

\begin{figure}
\centering
\includegraphics[width=0.9\linewidth]{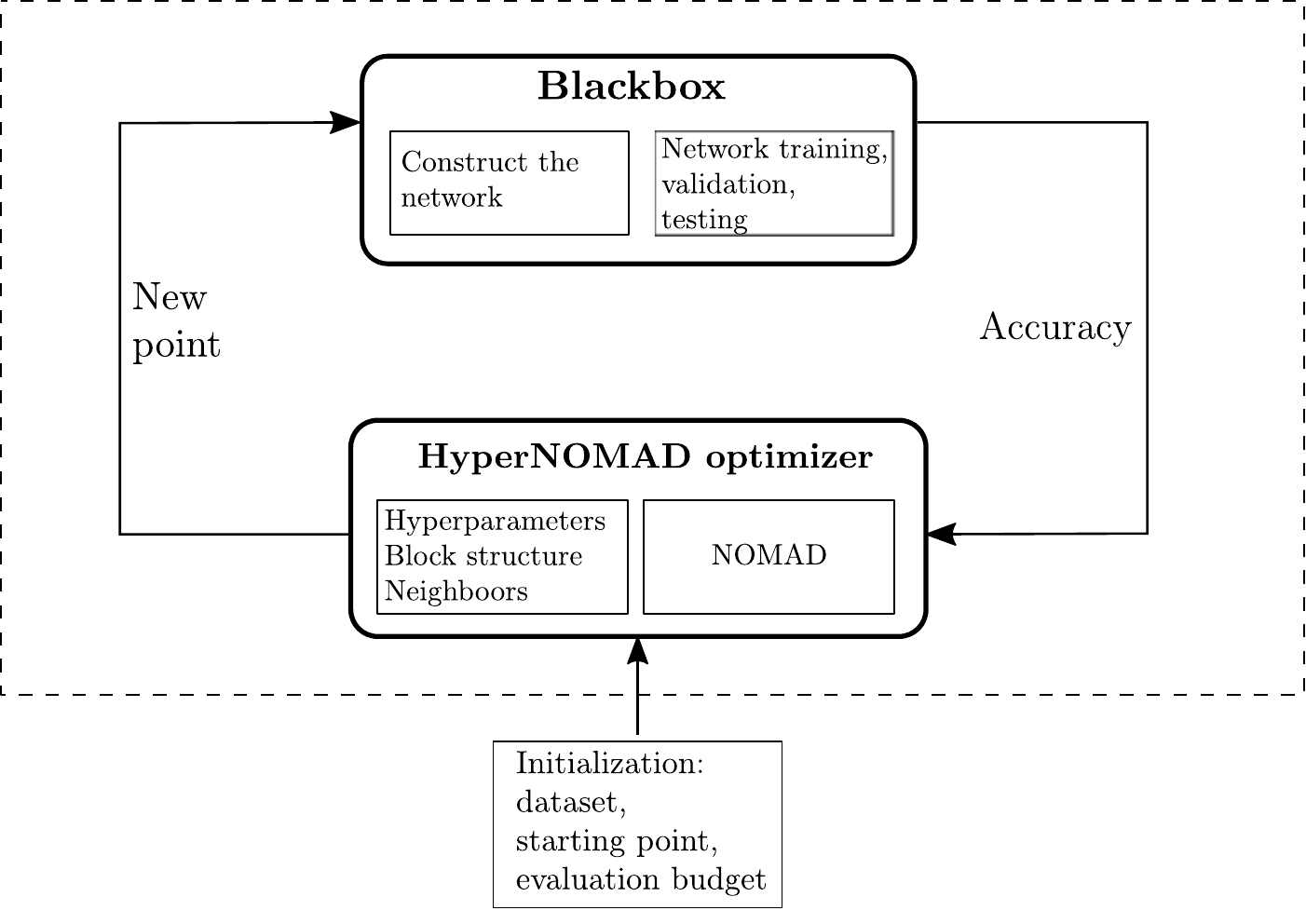}
\caption{The \hypernomad workflow.}
\label{fig-chart}
\end{figure}

This work presents \hypernomad, a package that applies \mads, a direct search method behind the \nomad software, to tune the hyperparameters that affect the architecture and the learning process of a deep neural network.   % The  chart 
Figure~\ref{fig-chart} illustrates the workflow when solving HPO problems with \hypernomad.
For a given set of hyperparameters, the construction of the network, the network training, validation and testing, are all wrapped as a single blackbox evaluation. One specificity of \hypernomad is its ability to explore a large search space by exploiting categorical variables.

The manuscript is structured as follows. Section~\ref{sec-lit} presents and discusses some of the main approaches used to solve the HPO problem of neural networks.
In Section~\ref{sec:exp}, the experimental setup is explicitly defined, and the instances used to test the proposed approach are presented.
Section~\ref{sec:hypernomad} introduces the \hypernomad package and gives an overview of  \mads, the algorithm selected to carry out the optimization task including the handling of categorical variables. Computational results are provided and discussed in Section~\ref{sec-results}.
Finally, Appendix~\ref{sec-appendix} describes the basic usage of \hypernomad.

%------------------------------------------------%
\section{Literature review}
\label{sec-lit}
%------------------------------------------------%

Tuning the hyperparameters of a deep neural network is a critical and time consuming process that was mainly done manually relying on the knowledge of the experts. However, the rising popularity of deep neural networks and their usage for diverse applications called for the automatization of this process in order to adapt to each problematic.

The hyperparameters that define a deep neural network can be separated into two categories: The ones that define the architecture of the network and the ones that affect the optimization process of the training phase. Tuning the hyperparameters of the first category alone has led to a separate field of research called Neural Architecture Search (NAS)~\cite{elsken2018neural} that allowed to achieve state of the art performances~\cite{real2018regularized, zoph2016neural} on some benchmark problems, although at a massive computational cost of 800 GPUs for a few weeks. Typically, one would perform a NAS first and then start tuning the other hyperparameters with the optimized architecture.
 However, Zela et al.~\cite{zela2018towards} argue that this separation is not optimal since the two aspects are not entirely independent from one another. Therefore, the proposed research considers both aspects at once.

One of the first scientific approach used to tackle the HPO problem of neural networks is the grid search. This method consists of discretizing the hypercube defined by the range of each hyperparameter and then evaluating each points on the grid. This technique is still used today and is implemented in several HPO libraries such as {\sf scikit-learn}  and {\sf Spearmint}~\cite {scikit-learn, Bayesian_HPO}. It has the advantage of being easy to understand, implement and parallelize. However, it becomes very expensive when training large networks, which is the case of deep neural networks, or when one seeks to optimize several hyperparameters at once. In addition, the grid search ignores the impact of each hyperparameter on the overall performance of the network. 

To avoid the drawbacks of the grid search, an alternative is to use random search~\cite{bergstra2012random}. Indeed, a random exploration of the space allows to evaluate more different values for each of the hyperparameters. This has the advantage of increasing the chances of finding a better configuration, but also to highlight the importance of some hyperparameters compared to the others. In addition, the random search makes it possible to highlight these properties with fewer evaluations than an exhaustive grid search. More recently, the {\sf Hyperband} algorithm~\cite{li2017hyperband} was introduced, which is a variant of a random search that uses an early stopping criteria to detect a non promising point early on in order to save computational resources and time. Thus achieving an important speedup compared to other methods. However, despite its advantages over the grid search, a random approach is limited because it is not adaptive and it does not exploit the performance scores of each configuration to direct the search. This can also waste resources that could have been better exploited by another optimization approach.

Genetic algorithms are evolutionary heuristics that are also used for the HPO problem. Inspired by biology, a genetic algorithm generates an initial population, i.e. a set of configurations, then, it combines the best parents to create a new generation of children. It also introduces random mutations to ensure a certain diversity in the population. These heuristics are therefore adaptive, thus allowing to explore the space more wisely even if they remain impregnated with a random character. These algorithms are often used to optimize hyperparameters~\cite{elsken2018efficient, suganuma2017genetic, young2015optimizing}. In~\cite {lorenzo2017particle}, a method based on particle swarm optimization is able to provide networks with higher performance than those defined by experts in less time than what would have required a grid search or a completely random search. Another approach using the evolutionary algorithm CMA-ES~\cite{loshchilov2016cma} was proposed with satisfactory results.

Other approaches based on machine learning can be found in the literature. For example, the HPO problem can be seen as reinforcement learning~\cite{baker2016designing, zoph2016neural, zoph2017learning} where the main difference between each method relies on how the agents are defined and dealt with. In~\cite{smithson2016neural}, a neural network is able to design other neural networks by learning to explore the possible configurations. Here, the HPO of neural networks is seen as a multiobjective problem where one seeks to improve the performance of the network while minimizing the computing power required. This approach, although successful, solves a different problem from the one considered in the context of this study. Also,~\cite{bosc2016} uses a network of long-term memory neurones to learn the parameters of another multi-layer network that is tested on a binary classification problem.

Derivative-Free Optimization (DFO) is naturally adapted to the HPO problem since it aims at solving problems typically given in the form of blackboxes that can be computationally costly to evaluate, with nonexistent or inoperable derivatives. In~\cite{AuOr06a}, the authors propose a general way of modeling hyperparameter optimization problems as a blackbox optimization problem. This formulation is used in~\cite {Liu2018}  to optimize 11 hyperparameters (3 real and 8  integer) of the BARON solver. This study compared the solutions found by  27 DFO algorithms on a total of 126  problems. The results show that the DFO methods have reduced the average resolution time, sometimes by more than 50\%. Another formulation inspired by robust optimization is used in~\cite{PoTo2017}, in addition to that of~\cite{AuOr06a}, to optimize the hyperparameters of the BFO algorithm~\cite{PoTo2017}. BBO methods are also at the heart of \vizier~\cite{golovin2017google} which is a tool that can be used for the HPO problem of machine learning algorithms, and especially for deep neural networks.

Bayesian optimization (BO) can be seen as a subclass of DFO methods and as such, can be used to solve the HPO problem. The BO methods use informations collected during previous assessments to diagnose the search space and predict which areas to explore first. Among them, Gaussian processes (GP) are models that seek to explain the collected observations that supposedly come from a stochastic function. GPs are a generalization of multi-variate Gaussian distributions, defined by a mean and a covariance function. GPs are popular models for optimizing the hyperparameters of neural networks~\cite {Bayesian_HPO, wistuba2018scalable}. However, the disadvantage of GPs is that they do not fit well to the categorical features, and its performance depends on the choice of the kernel function that defines it.  Tree-structured Parzen Estimator (TPE) is also a Bayesian method that can be used as a model instead of a GP. After a certain number of evaluations, this method separates the evaluated points into two sections: a portion  ($<$25\%) of the points with the best performances, and the remaining. This method seeks to find a distribution of the best observations to determine the next candidates.TPEs are also used for the HPO of neural networks~\cite{bergstra2011algorithms}, even if it has the disadvantage of ignoring the interactions between the hyperparameters.

Other model-based DFO methods were also applied to the HPO problem. In~\cite{Nannicini_HPO}, the authors applied radial basis functions to model the blackbox as previously defined. This article presents the results obtained on the MNIST data set~\cite{mnist}, then on a problem of interactions between drugs. These tests show that this model provides comparable or better results than popular configurations. In~\cite{ghsc2017}, a trust-region DFO algorithm is applied to optimize the hyperparameters of a SVM model. Here again, this approach obtained a more efficient model than those defined by the experts or by a Bayesian algorithm.

Thus, the positive results of these methods suggest that the DFO approach is well suited to solve the HPO problem of deep neural networks. This motivated the idea of using the \mads algorithm~\cite{AuDe2006}, which is implemented into the \nomad package~\cite{Le09b}, especially as it can handle integer and categorical variables~\cite{AACW09a, AuLeDTr2018}.
Using the \mads algorithm for hyperparameter tuning
has been  validated in~\cite{mello2019novel}
where a SVM model is calibrated using MADS combined with
the Nelder-Mead and VNS search strategies~\cite{AuBeLe08,AuTr2018}.

A non exhaustive list of open source librairies for HPO is given in Table~\ref{tab:lib_hpo} along with the optimization algorithms implemented in each library and the types of variables handled.

\begin{table}[ht]
  \caption{Selection of open source libraries for the hyperparameter optimization problem.}
    \centering
   \begin{tabular}{l c c c c c c c r} \toprule
   	{  } & \multicolumn{5}{c }{Optimization method} & \multicolumn{3}{c }{Type of variables} \\ \midrule
	 Package & \makecell{Grid \\ search}  &  \makecell{Random\\ search} &  \makecell{Bayesian \\ optimization} &  \makecell{Model- \\ based} &  \makecell{Direct- \\ search} & Real & Int. & Cat. \\ \midrule
	 {\sf scikit-learn}~\cite{scikit-learn} &  \cmark  &  \cmark  &   - & - &  -  & \cmark & \cmark &  \cmark  \\
	\hyperopt~\cite{bergstra2013making} &   -   & \cmark & \cmark &  -  & -   & \cmark & \cmark & \cmark \\
	{\sf Spearmint}~\cite{Bayesian_HPO} & \cmark & \cmark & \cmark & -   &   -  & \cmark & \cmark & \cmark \\
	{\sf SMAC}~\cite{hutter2011sequential} &  -  &  -  &  -  & \cmark &  -  & \cmark & \cmark & \cmark \\
	{\sf MOE}~\cite{moe}&  -  &   - & \cmark &  -  &  -   & \cmark &  -  &  -  \\
	{\sf RBFOpt }~\cite{Nannicini_HPO} &  -  &  -  &   - & \cmark &  -  & \cmark & \cmark &  -  \\
	{\sf DeepHyper}~\cite{deephyper} &  -  &  \cmark  &   - & \cmark &  -  & \cmark & \cmark &  \cmark  \\
	{\sf Or\'ion}~\cite{orion} &  -  &  \cmark  &   - & - &  -  & \cmark & \cmark &  \cmark  \\
	\vizier~\cite{golovin2017google} & \cmark & \cmark & \cmark & \cmark & - & \cmark & \cmark & \cmark \\
	\hypernomad  & -   &  -  &   - &  -  & \cmark & \cmark & \cmark & \cmark \\  \bottomrule
   \end{tabular}
   \label{tab:lib_hpo}
\end{table}

%--------------------------------------------------------------%
\section{Experimental setup} \label{sec:exp}
%--------------------------------------------------------------%

This section first defines the blackbox approach used for modeling the HPO problem. This is done by listing the different hyperparameters considered to construct, train and validate a deep neural network (DNN) in order to obtain its test accuracy. The second part of the section gives an overview of the data sets provided with \hypernomad.

%--------------------------------------------------------------%
\subsection{Hyperparameters of the framework}
\label{sec:hpos}
%--------------------------------------------------------------%

A variety of hyperparameters must be chosen to tune a DNN for a given application. These hyperparameters affect different aspects of the network: the architecture, the optimization process and the handling of the data. The following section lists the hyperparameters considered in this study along with their respective types and scopes.

\subsubsection{The network architecture}

\begin{figure}[ht]
   \centering
   \begin{tabular}{@{}c@{\hspace{.5cm}}c@{}}
      \includegraphics[scale=0.65]{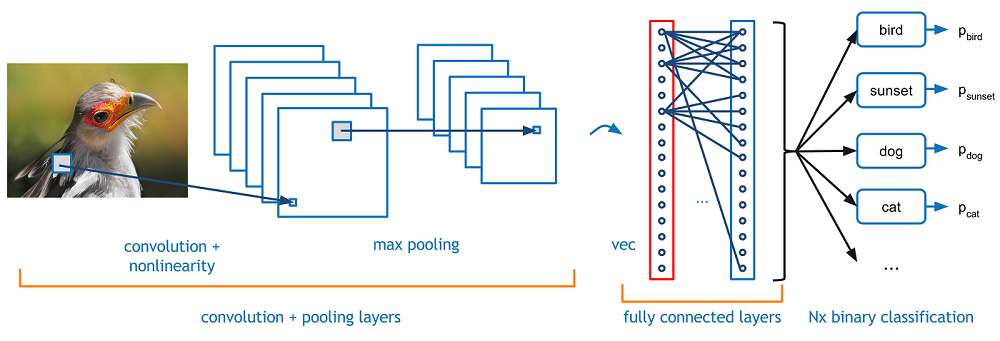} \\
   \end{tabular}
 \caption{Example of a convolutional neural network. Image taken from~\cite{schema_cnn}.}
 \label{fig:convnet}
\end{figure}

A convolutional neural network (CNN) is a deep neural network consisting of a succession of convolutional layers followed by fully connected layers as illustrated in Figure~\ref{fig:convnet}.

\begin{figure}[ht]
   \centering
   \begin{subfigure}{0.52\textwidth}
	\includegraphics[width=\linewidth]{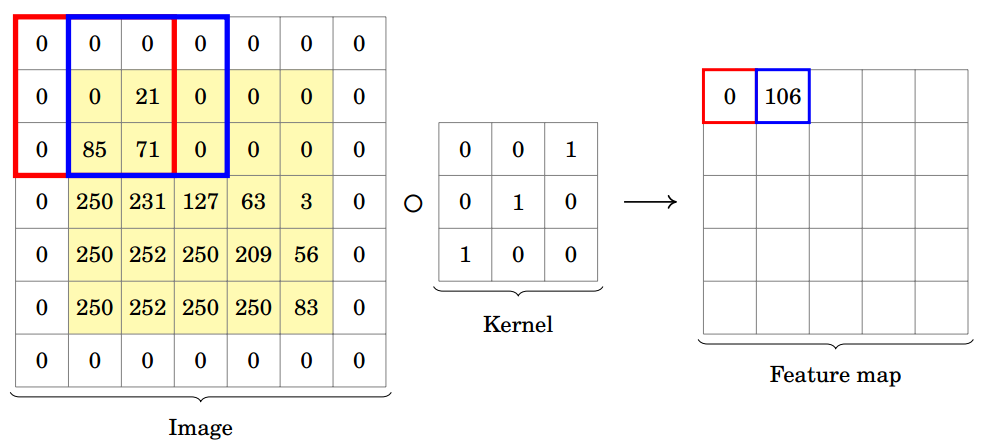}
	\caption{Convolution} \label{fig:1a}
   \end{subfigure}
   \hspace*{\fill} % separation between the subfigures
   \begin{subfigure}{0.45\textwidth}
	\includegraphics[width=\linewidth]{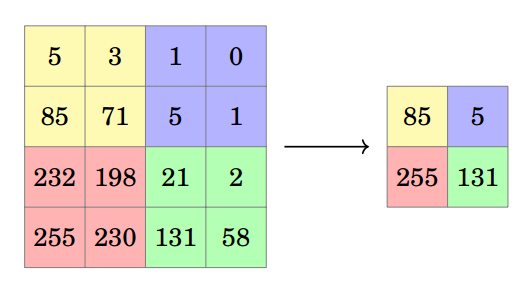}
	\caption{Pooling} \label{fig:1b}
   \end{subfigure}
 \caption{Illustration of a convolution operation in (a) and a pooling operation in (b). Images taken from~\cite{schema_cnn2}.}
 \label{fig:convolution}
\end{figure}

To define a new CNN,  one must first decide on the number of convolutional layers. These layers can be seen as matrices in a two dimensional convolution. The size of the first convolutional layer is determined by the size of the images the network is fed. The size of the remaining layers is computed by taking into account the different operations applied from layer to layer. These operations can be divided into two categories: a convolution or a pooling. Figure~\ref{fig:1a} represents the steps of a convolution operation. The initial image is a 5$\times$5 matrix whose borders are padded with zeros. The convolution consists of choosing a kernel that is passed over the image in order to compute the coefficients of the feature map. Each coefficient is equal to the sum of the products between the coefficients of the image and the ones of the kernel situated in the same position. In Figure~\ref{fig:1a}, the coefficient (2,1) of the feature map is obtained by the following operation: $(0\times0) + (0\times0) + (0\times1) + (0\times0) + (21\times1) + (0\times0) + (85\times1) + (71\times0) + (0\times0) = 106$.
In general, a convolution can be determined with few factors such as the number of feature maps - or output channels - generated, the size of the kernel which in turn will affect the size of the feature map, the stride which corresponds to the step by which the kernel is moved over the image and the padding. In Figure~\ref{fig:1a}, the image is padded with one layer of zeros.

When the feature map is obtained, one can decide to apply a pooling operation to decrease the size of the output by keeping only the biggest coefficients in a certain area. Figure~\ref{fig:1b} illustrates a 2$\times$2 pooling that results in an output of half the size of the feature map.

\begin{figure}[ht]
   \centering
    \begin{subfigure}{0.32\textwidth}
	\includegraphics[width=\linewidth]{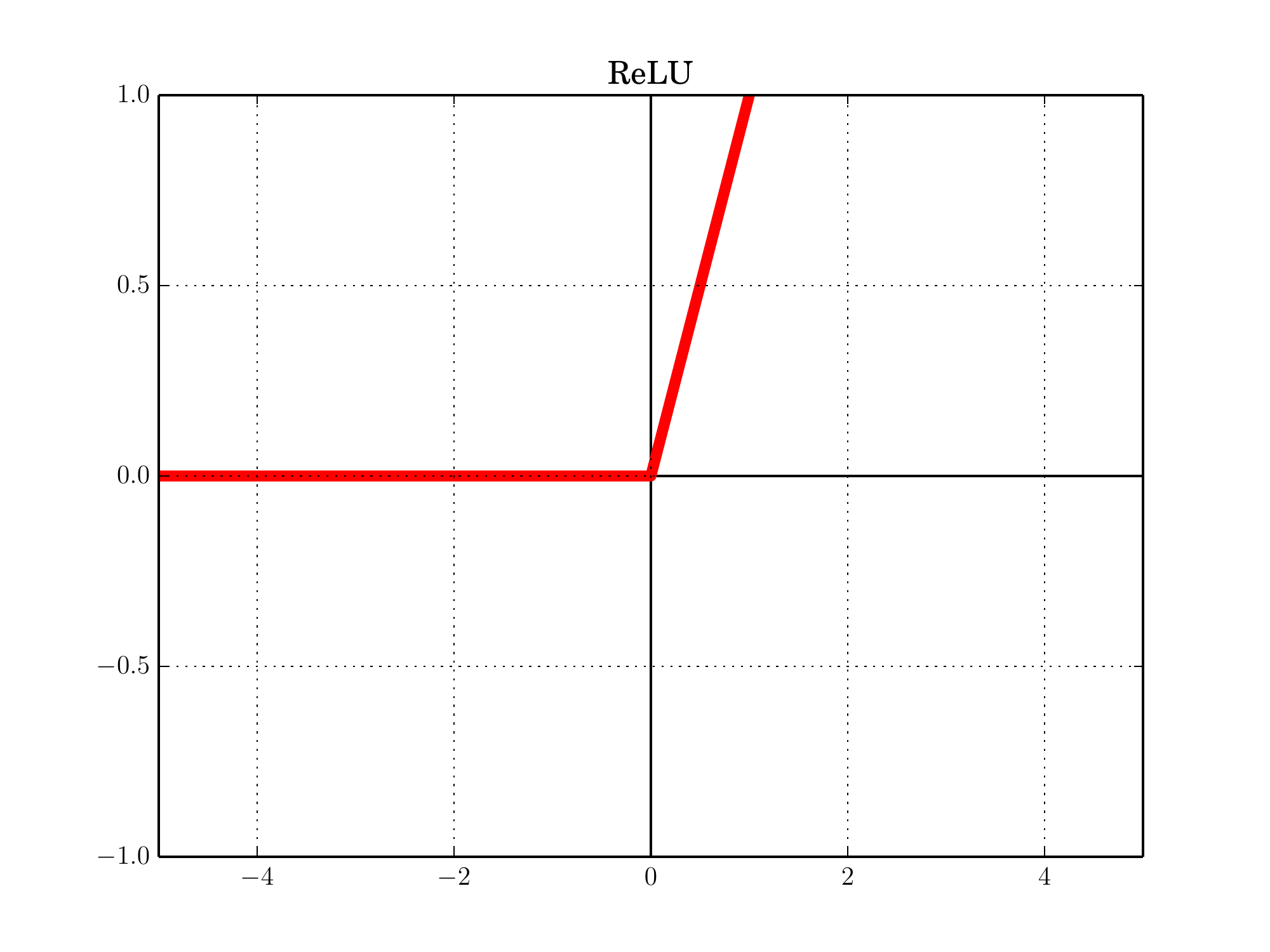}
	
   \end{subfigure}
   \hspace*{\fill} % separation between the subfigures
   \begin{subfigure}{0.32\textwidth}
	\includegraphics[width=\linewidth]{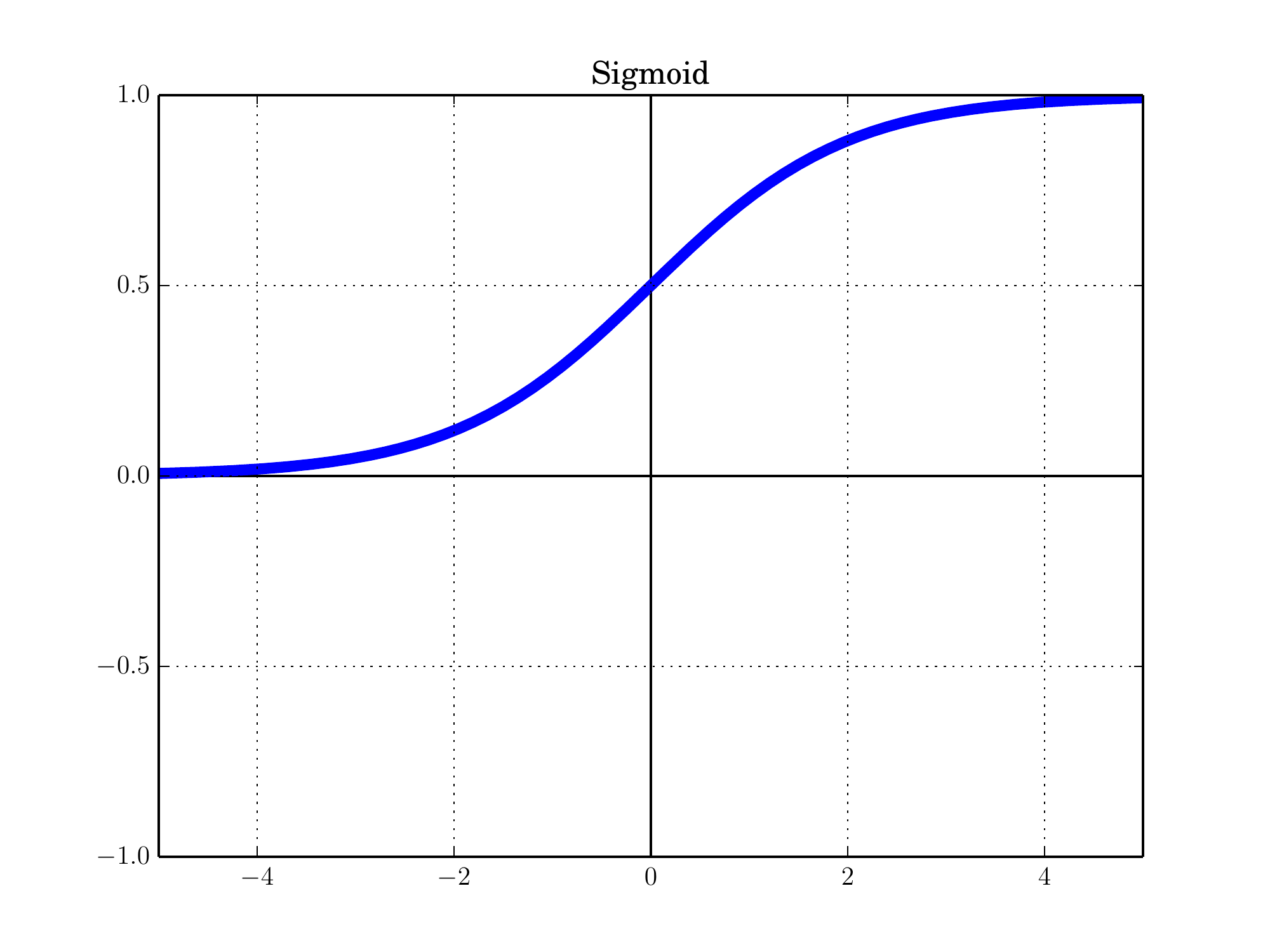}
   \end{subfigure}
    \begin{subfigure}{0.32\textwidth}
	\includegraphics[width=\linewidth]{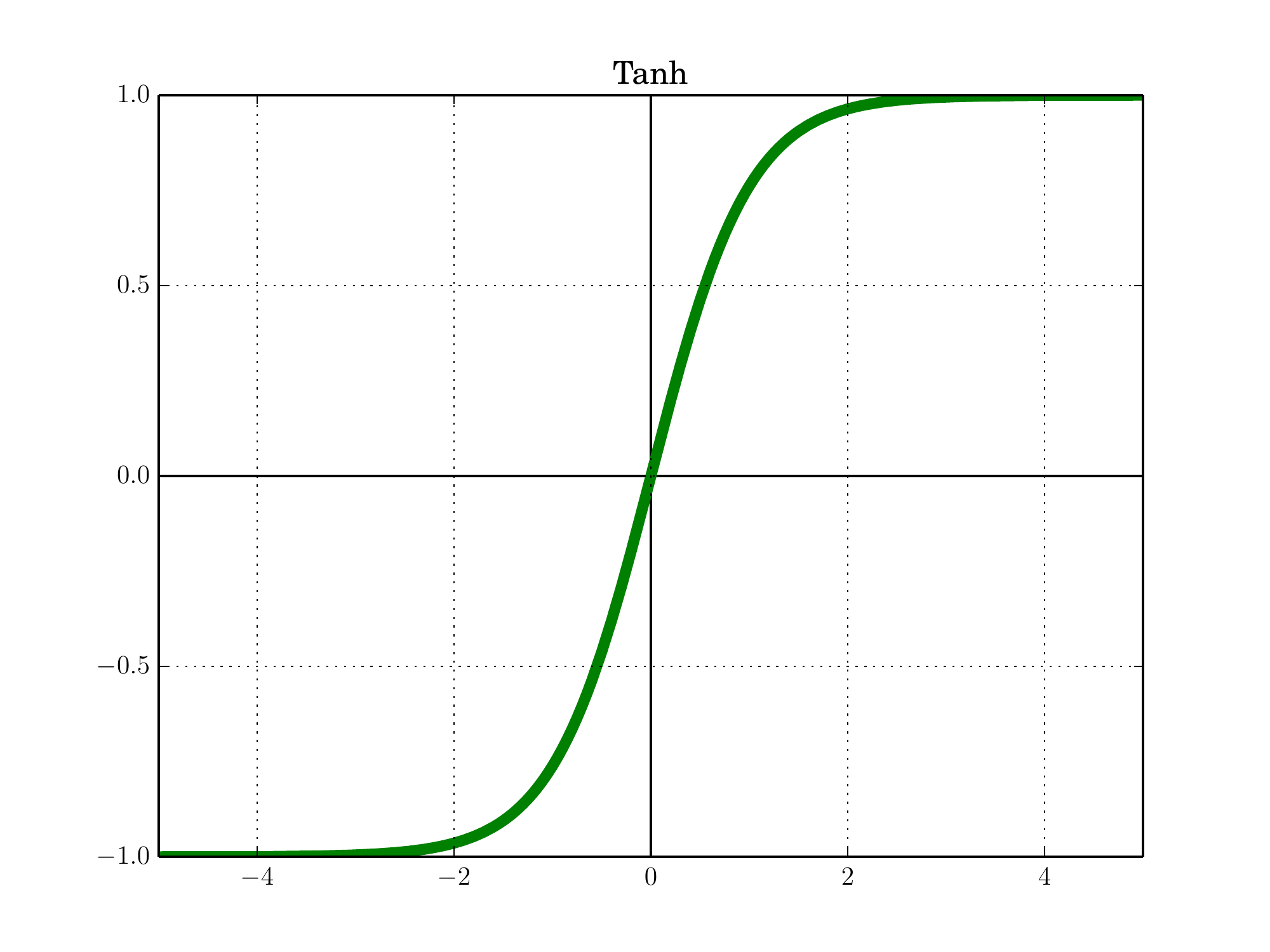}
   \end{subfigure}  
 \caption{Examples of activation functions.}
 \label{fig:activation}
 \end{figure}

Each fully connected layer that follows the convolutional ones is determined by the number of neurones it contains.  The neurones of a layer are connected to all of the ones in the next layer through weighted arcs. Let $x_1, x_2, \ldots, x_{n_l}$ be the values of the neurones of the layer $l$ and $a_j$ be the value of the neurone $j$ in the layer $l+1$, then $ a_j = \phi(\sum\limits_{i =1}^{s_l} w_{ij} x_i)$, where $w_{1j}, w_{2j}, \ldots, w_{{n_l}j}$ are the weights of the arcs from the neurones of the layer $l$ to the $j-th$ neurone of the following layer and $\phi$ is an activation function used to introduce a non linearity in the outputs. Figure~\ref{fig:activation} presents some examples of activation functions.

Table~\ref{tab:HPNN} summarizes the hyperparameters responsible for defining the structure of the network. 
Hyperparameters 2 to 6 must be defined for each convolutional layer and the hyperparameter 8 must also be defined for each fully connected layer. Therefore, if $n_1$ is the number of convolutional layers and $n_2$ the number of fully connected layers, the total number of hyperparameters responsible for defining the structure of the neural network is $5n_1 + n_2 + 4$.

\begin{table}[ht]
  \caption{Hyperparameters that define the architecture of a neural network.}
    \centering
  %  |lllr
   \begin{tabular}{p{0.08 \textwidth} p{0.45 \textwidth} p{0.2 \textwidth} p{0.22 \textwidth}  }  \toprule
   	% \hline
   	 {\#} & \textbf{Hyperparameter} & \textbf{Type} &  \hspace*{1cm}\textbf{Scope} \\ % \midrule
	\hline
	1 & Number of convolutional layers $(n_1)$& Categorical &  \hspace*{1cm}$\{$0,1,\ldots,20$\}$\\
	2 & \hspace*{5mm}Number of output channels & Integer &  \hspace*{1cm}$\{$0,1,\ldots,50$\}$\\
	3 & \hspace*{5mm}Kernel size & Integer &  \hspace*{1cm}$\{$0,1,\ldots,10$\}$\\
	4 & \hspace*{5mm}Stride & Integer &  \hspace*{1cm}$\{$1,2,3$\}$ \\
	5 & \hspace*{5mm}Padding & Integer &  \hspace*{1cm}$\{$0,1,2$\}$\\
	6 & \hspace*{5mm}Do a pooling & Boolean &  \hspace*{.9cm} $\{$0,1$\}$ \\
	% \hline
	7 & Number of full layers $(n_2)$& Categorical &  \hspace*{1cm}$\{$0,1,\ldots,30$\}$\\
	8 & \hspace*{5mm}Size of the full layer & Integer &  \hspace*{1cm}$\{$0,1,\ldots,500$\}$\\ 
	% \hline
	9 & Dropout rate & Real &  \hspace*{1cm}[0;1] \\ 
	% \hline
	10 & Activation function & Categorical/Integer &  \makecell{$\{$ReLU (1), \\ Sigmoid (2), \\ Tanh (3)$\}$}  \\ 
	\bottomrule
   \end{tabular}
   \label{tab:HPNN}
\end{table}

%--------------------------------------------------------------%
\subsubsection{The optimizer}
%--------------------------------------------------------------%

For a given network architecture, the training phase is conducted to minimize
the error between the predictions of the network and the correct values of the labels attached to the validation data. Let $\Theta$ be a multi-dimensional matrix that stores the weights of the arcs that link each layer of the network $l$ with the next one $l+1$, and let $J(\Theta)$ the sum of the errors between the predictions and the labels for all the data. The optimizer must then solve $\min_{\Theta } J(\Theta).$  Before starting the training phase, the optimizer that carries out this task must be selected along with its specific algorithmic hyperparameters.

A stochastic gradient approach is more suitable in this case because of the high dimension of this problem which is usually in the millions. At each iteration, the weights of the network are updated by following a stochastic direction with a particular step size which is called a learning rate in the machine learning context. Similarly to any gradient descent method, the learning rate must be chosen and updated accordingly to avoid oscillations or divergence. Substantial research and tricks of the trade are developed to solve this problematic~\cite{bengio2012practical, stochastic-gradient-tricks, LeCun2012}. The optimizers Adam~\cite{Adam}, Adagrad~\cite{duchi2011adaptive} and RMSProp~\cite{rmsprop} have embedded strategies to adapt the learning rate at each iteration and for each weight. SGD however does not require external management. In \hypernomad, the learning rate of SGD is divided by 10 every 100 epochs as long as its value is greater than $10^{-6}$.

Table~\ref{tab:optimizer} presents the list of selectable optimizers considered in the blackbox along with their corresponding hyperparameters. There is one categorical hyperparameter that determines which optimizer is chosen and always four real hyperparameters related to it. This aspect of the network relies on defining five hyperparameters in total.

\begin{table}[ht]
\caption{Choices of the optimizer and the corresponding hyperparameters.}
  \centering
   \begin{tabular}{p{0.5 \textwidth}  p{0.25 \textwidth}  p{0.1 \textwidth}  p{0.1 \textwidth}} \toprule
   	\textbf{Optimizer } & \textbf{Hyperparameter} & \textbf{Type} & \textbf{Scope} \\ \midrule
	Stochastic Gradient Descent (SGD)	& Initial learning rate 		& Real 	& [0;1]\\
	 						       	& Momentum 			& Real 	& [0;1]\\
							       	& Dampening 			& Real 	& [0;1]\\
							       	& Weight decay 		& Real 	& [0;1]\\ \midrule
	Adam						& Initial learning rate 		& Real 	& [0;1]\\
								& $\beta_1$			& Real 	& [0;1]\\
								& $\beta_2$			& Real 	& [0;1]\\
								& Weight decay 		& Real 	& [0;1]\\ \midrule
	Adagrad						& Initial learning rate 		& Real 	& [0;1]\\
								& Learning rate decay	& Real 	& [0;1]\\
								& Initial accumulator		& Real 	& [0;1]\\
								& Weight decay 		& Real 	& [0;1]\\ \midrule
	RMSProp						& Initial learning rate 		& Real 	& [0;1]\\
								& Momentum 			& Real 	& [0;1]\\
								& $\alpha$			& Real 	& [0;1]\\   
								& Weight decay 		& Real 	& [0;1]\\					       
	 \bottomrule
   \end{tabular}
   \label{tab:optimizer}
\end{table}

\subsubsection{The training phase}
Before training a network, the data must be separated into three groups, each one is responsible for the training, the validation and the testing of the network. During the training phase, the network is fed with the training data, performs a forward pass, computes the prediction error and do a back-propagation in order to update the weights using the optimizer. The way the network is fed is also of great importance. One can choose to input the training data one by one, all at once, or by sending subsets or mini-batches of the data. The size of the mini-batches is an integer hyperparameter that varies between $[1, n_{train}]$, where $n_{train}$ is the size of the training data.

When the network has been fed all of the training data, it is said to have performed an epoch. Usually, the training data has to be passed more than once in order to obtain good weights and a good testing accuracy. Therefore, the number of epochs must be chosen as well. This hyperparameter is dealt with as follows. The validation accuracy is evaluated after each epoch and the weights of the network responsible for the best validation accuracy are stored. This process is repeated as long as the number of epochs is lower than a certain maximum number of epochs (usually 500) and as long as an early stopping condition has not been satisfied. These early stopping criteria depend on the evolution of the training and validation of the network. When the validation accuracy staggers or when it stays lower than 20\% after 50 epochs then the training can be interrupted in order to save time and computational ressources. Once the training is done, the test accuracy is evaluated using the weights that gave the best validation accuracy.

Finally, the blackbox optimization problem is obtained following the model in~\cite{AuOr06a}. This blackbox takes $5n_1 + n_2 + 10$ mixed variable inputs, where $n_1$ is the number of convolutional layers and $n_2$ the number of fully connected layers of the network, and returns the value of the accuracy on the test data set. This blackbox problem is solved using the \nomad software~\cite{Le09b} described in Section~\ref{sec:nomad}. 

%------------------------------------------------%
\subsection{Data sets}
%------------------------------------------------%

The \hypernomad package comes with a selection of data sets all meant for classification problems. Table~\ref{tab:datasets} lists the data sets embedded so far through \pytorch~\cite{paszke2017automatic}, a relatively complete tool to model and manipulate deep neural networks. \hypernomad also allows the usage of a personal data set by following the instructions given in Appendix~\ref{sec-appendix}.
When loading a data set from Table~\ref{tab:datasets}, \hypernomad applies a normalization and a random horizontal flip to regulate and augment the data.

\begin{table}[ht]
\caption{Data sets embedded in \hypernomad.}
  \centering
   \begin{tabular}{l r r r r} \toprule
%     \begin{tabular}{p{0.2 \textwidth} p{0.2 \textwidth} p{0.2 \textwidth} p{0.15 \textwidth} p{0.15 \textwidth}} \toprule
   \textbf{Data set } & \textbf{\makecell{Training \\ data}} & \textbf{ \makecell{Validation\\ data}} &\textbf{\makecell{Testing \\data}} & \textbf{\makecell{Number \\ of classes}} \\ \midrule
   MNIST 	   			&  \hspace*{1cm}40,000 	     & \hspace*{1cm}10,000	          & \hspace*{1cm}10,000   & \hspace*{1.25cm}10\\
   Fashion-MNIST		& \hspace*{1cm}40,000 	     & \hspace*{1cm}10,000	          & \hspace*{1cm}10,000	   & \hspace*{1.25cm}10\\
   EMNIST 	   		&  \hspace*{1cm}40,000 	     & \hspace*{1cm}10,000	          & \hspace*{1cm}10,000	   & \hspace*{1.25cm}10\\
   KMNIST 	   		&  \hspace*{1cm}40,000 	     & \hspace*{1cm}10,000	          & \hspace*{1cm}10,000	   & \hspace*{1.25cm}10\\
   CIFAR-10	   		&  \hspace*{1cm}40,000 	     & \hspace*{1cm}10,000	          & \hspace*{1cm}10,000	   & \hspace*{1.25cm}10\\
   CIFAR-100 	   		&  \hspace*{1cm}40,000 	     & \hspace*{1cm}10,000	          & \hspace*{1cm}10,000	   &\hspace*{1.25cm}100\\
   STL-10 	   			&  \hspace*{1cm}4,000 	     & \hspace*{1cm}1,000	          & \hspace*{1cm}8,000	   &\hspace*{1.25cm}10\\
   \bottomrule
   \end{tabular}
   \label{tab:datasets}
\end{table}

The rest of the section describes the data sets used for benchmarking \hypernomad. First, a validation is done using MNIST~\cite{mnist} and once positive results are obtained, the second and more complex data set, CIFAR-10~\cite{krizhevsky2009learning}, is considered.

%------------------------------------------------%
\subsubsection{MNIST}\label{sec:dataset_mnist} %  / Caffe
%------------------------------------------------%

MNIST~\cite{mnist} is a data set containing 60,000 images of hand written digits that is usually divided into three categories: 40,000 for training, 10,000 for validation and the remaining 10,000 for testing. The set is used for developing a convolutional neural network capable of recognizing the digits in each image and assigning it to the correct class. 
The relative simplicity of this task does not require complex neural networks to obtain a good accuracy. Therefore, this data set is usually considered as a first validation of a concept and not a sufficient proof of the quality of a method among the machine learning community.

%------------------------------------------------%
\subsubsection{CIFAR-10} 
%------------------------------------------------%

The second set of tests are performed with CIFAR-10~\cite{krizhevsky2009learning}. This data set contains 60,000 colored images of objects that belong to ten different and independent categories. The data is once again divided into three sets: 40,000 for training, 10,000 for validation and 10,000 for testing.

For this test, the blackbox within \hypernomad is used to construct the convolutional neural network corresponding to the values of the hyperparameters described in Section~\ref{sec:hpos}. This network is trained, validated and tested on CIFAR-10 according to the mode of operation of \hypernomad detailed in Section~\ref{sec:hypernomad}.

%------------------------------------------------%
\section[HyperNOMAD]{\hypernomad}\label{sec:hypernomad}
%------------------------------------------------%

The  \hypernomad package is available on GitHub\footnote{\url{\hypernomadurl}}.
It contains a series of {\sf Python} programs wrapped into a blackbox responsible for constructing, training and evaluating the test accuracy of a neural network depending on the values of the hyperparameters described in Section~\ref{sec:exp}. This blackbox uses the \pytorch package~\cite{paszke2017automatic}  for its simplicity. \hypernomad also contains an interface that runs the optimization of the blackbox using the \nomad software~\cite{Le09b} described in the rest of this section. The basic usage of  \hypernomad is described in Appendix~\ref{sec-appendix}.

%------------------------------------------------%
\subsection{Overview of \nomad}\label{sec:nomad}
%------------------------------------------------%

The \nomad software~\cite{Le09b} is a {\sf C++} implementation of the \mads algorithm~\cite{AuDe2006,AuLeDTr2018} which is a direct search method that generates, at each iteration $k$, a set of points on the {\em mesh}
$M^k = \left\{ x +  \diag(\deltA^k) z : x \in V^k, z \in \Z^n \right\}$
where $V^k$ contains the points that were previously evaluated (including the current iterate $x^k$)
and  $\deltA^k \in \R^n$ is the {\em mesh size vector}.

Each iteration of \mads is divided into two steps: The \textit{search} and the \textit{poll}. The \textit{search} phase is optional and can contain different strategies to explore a wider space in order to generate a finite number of possible mesh candidates. This step can be based on surrogate functions, latin hyper-cube sampling, etc. The \textit{poll}, on the other hand, is strictly defined since the convergence theory of \mads relies solely on this phase. Here, the algorithm generates directions around the current iterate $x^k$ to search for candidates locally in a region centered around  $x^k$ and of radius, in each dimension, of $\DeltA^k \in \R^n$, which is called the {\em poll size vector}. The set of candidates in this phase defines the {\em poll set} $P_k$.

If \mads finds a better point then the iteration is declared a success and the mesh and poll sizes are increased, however, if the iteration fails then both parameters are reduced so that
$\deltA^k \leq \DeltA^k$ is maintained.  This relation insures that the set of search directions becomes dense in the unit sphere asymptotically.
The \mads algorithm is summarized in Algorithm~\ref{algo-mads}.

\begin{algorithm}
\caption{Mesh adaptive direct search (\mads)}
\label{algo:mads}
\begin{flushleft}
$k = 0, \deltA^0$, $x^0$

\textbf{[1] Search (optional)}\\
\hspace*{10mm} Construct a set of mesh points and evaluate them \\
\hspace*{10mm} If there is a success, go to \textbf{[3]}

\textbf{[2] Poll}\\
\hspace*{10mm} Evaluate the points in the poll set $P_k$

\textbf{[3] Updates}\\
\hspace*{10mm} Update $\deltA^k, x^k, M^k, V^k$ depending on the success of the previous phases\\
\hspace*{10mm} If no stopping condition is satisfied: $k \leftarrow k+1$ and go to \textbf{[1]}
\end{flushleft}
\label{algo-mads}
\end{algorithm}

In addition, \nomad can handle categorical variables by adding a step in the basic \mads algorithm.
A variable is categorical when it can take a finite number of nominal or numerical values that express a qualitative property that assign the variable to a class (or category). The algorithm relies on an ad~hoc neighborhood structure, provided in practice by the user as a list of neighbors  for any given point. The poll step of \mads is augmented with the so-called \textit{extended poll} that links the current iterate $x^k$ with the independent search spaces where the neighbors can be found. The first neighbor that improves the objective function is chosen and the optimization carries on in the corresponding search space.
For more detail on how \mads handles categorical variables, the reader is referred to the following list of articles~\cite{Abra04,AACW09a,AbAuDe2007a,AuDe01a,KoAuDe01a}.

%------------------------------------------------------------%
\subsection[Hyperparameters in HyperNOMAD]{Hyperparameters in \hypernomad}
%------------------------------------------------------------%

The selected neighborhood structure in \hypernomad relies on blocks of categorical variables with their associated variables.
The following subsections describe this structure.

% The \hypernomad package stores and manipulates hyperparameters through block structures which rely on exploiting categorical variables. 
% The following sections describe these variables and illustrate the blocks in \hypernomad through an example.
% -------------------------------------------------
% \subsubsection{Categorical variables \label{sec:cat_var}}
% -------------------------------------------------
% {\bl One important property of \nomad is that it can handle mixed variable problems~\cite{AACW09a, AuLeDTr2018}. }

% A variable is categorical when it can take a finite number of nominal or numerical values that express a qualitative property that assign the variable to a class (or category). {\bl In \nomad, handling categorical variables requires to provide an ad~hoc neighborhood structure given as one or more neighbor points for any given point. In \hypernomad, the \textit{poll} step of \mads is supplied with an \textit{extended poll} that links the current best point (also called current iterate) with the independent search spaces where the neighbors can be found~\cite{AACW09a}. The first neighbor that improves the objective function is chosen and the optimization carries on in the corresponding search space.

%------------------------------------------------------------%
\subsubsection{Blocks of hyperparameters}
%------------------------------------------------------------%

 \hypernomad splits the hyperparameters (HPs) defined in Section~\ref{sec:hpos} into different blocks: one for the convolution layers, the fully connected layers, the optimizer and one for each of the other HPs. A block is an implemented structure that stores a list of values, each one starting with a header and followed by the associated variables, when applicable, that are gathered into groups.
For example, consider a CNN with two convolutional layers, each one defined with the number of output channels, the kernel size, the stride, the padding and whether a pooling is applied or not as stated in Table~\ref{tab:HPNN}. Then consider the values $(16, 5, 1, 1, 0)$ and $(7, 3, 1, 1, 1)$. Each set of values corresponds to a group of variables that describes one convolutional layer and both groups constitute the convolution block. The header of the convolution block is the categorical variable that represents the number of convolutional layers ($n_1$) that the CNN contains as shown in Figure~\ref{fig:conv_block} (top).
The convolution block is followed by the fully connected block. The header of this block also corresponds to the categorical variable that describes the number of fully connected layers. Here, each layer is defined with the number of neurones it contains. Therefore, if $n_2$ is the value in the header, then  there are $n_2$ groups of a single variable as illustrated in Figure~\ref{fig:full_block} (top).
The optimizer block always possesses a fixed size since there is always five HPs that describe the optimizer: The choice of the algorithm and four related HPs as summarized in Table~\ref{tab:optimizer}. The header of this block is the categorical variable corresponding to the choice of the optimizer and the four associated variables are gathered into one group as shown in Figure~\ref{fig:opt_block}.
The other HPs are put as the headers of their individual block with no associated variable.

\begin{figure}[!htb]
  \flushleft 
\hspace*{3.25em}\begin{tabular}{| p{5mm} | p{5mm} | p{5mm} | p{5mm} | p{5mm} | p{5mm} | p{5mm} | p{5mm} | p{5mm} | p{5mm} | p{5mm} |} 
   	\hline
   	\cellcolor{green!25}{2 } & \cellcolor{blue!25}{16} & \cellcolor{blue!25}{5} & \cellcolor{blue!25} {1} & \cellcolor{blue!25}{1} & \cellcolor{blue!25}{0} & \cellcolor{yellow!25}{7} & \cellcolor{yellow!25}{3} & \cellcolor{yellow!25} {1} & \cellcolor{yellow!25}{1} &\cellcolor{yellow!25} {1} \\
	\hline
   \end{tabular}
   \hspace*{3em}
   \begin{tabular}{| p{5mm} | p{5mm} | p{5mm} | p{5mm} | p{5mm} | p{5mm} | p{5mm} | p{5mm} | p{5mm} | p{5mm} | p{5mm} |p{5mm} | p{5mm} | p{5mm} | p{5mm} | p{5mm} |} 
   	\hline
   	\cellcolor{green!25}{3 } & \cellcolor{blue!25}{16} & \cellcolor{blue!25}{5} & \cellcolor{blue!25} {1} & \cellcolor{blue!25}{1} & \cellcolor{blue!25}{0} & \cellcolor{yellow!25}{7} & \cellcolor{yellow!25}{3} & \cellcolor{yellow!25} {1} & \cellcolor{yellow!25}{1} &\cellcolor{yellow!25} {1} & \cellcolor{red!25}{7} & \cellcolor{red!25}{3} & \cellcolor{red!25} {1} & \cellcolor{red!25}{1} &\cellcolor{red!25} {1} \\
	\hline
   \end{tabular}
   \hspace*{3em}
   \begin{tabular}{| p{5mm} | p{5mm} | p{5mm} | p{5mm} | p{5mm} | p{5mm} |} 
   	\hline
   	\cellcolor{green!25}{1} & \cellcolor{blue!25}{16} & \cellcolor{blue!25}{5} & \cellcolor{blue!25} {1} & \cellcolor{blue!25}{1} & \cellcolor{blue!25}{0}  \\
	\hline
\end{tabular}
      \caption{Example of a convolution block (top). Its first neighbor is obtained by adding a convolutional layer (middle) and the second neighbor is obtained by subtracting a convolutional layer (bottom).}
\label{fig:conv_block}
   \end{figure}

\begin{figure}[!htb]
       \flushleft 
\hspace{8em}\begin{tabular}{| p{5mm} | p{10mm} | p{10mm} | p{10mm} | } 
   		\hline
   		\cellcolor{green!25}{3 } & \cellcolor{blue!25}{1200} & \cellcolor{yellow!25}{512} & \cellcolor{red!25} {20}\\ 
		\hline
   	\end{tabular}

\hspace{8em}\begin{tabular}{| p{5mm} | p{10mm} | p{10mm} | p{10mm} | p{10mm} | } 
   		\hline
   		\cellcolor{green!25}{4 } & \cellcolor{gray!25}{1200}  & \cellcolor{blue!25}{1200} & \cellcolor{yellow!25}{512} & \cellcolor{red!25} {20}\\ 
		\hline
   	\end{tabular}

\hspace{8em}\begin{tabular}{| p{5mm} | p{10mm} | p{10mm} | } 
   		\hline
   		\cellcolor{green!25}{2 }& \cellcolor{yellow!25}{512} & \cellcolor{red!25} {20}\\ 
		\hline
   	\end{tabular}
		\caption{Example of a fully connected block (top). Its first neighbor is obtained by adding a fully connected layer (middle) and the second neighbor is obtained by subtracting a fully connected layer (bottom).}
		\label{fig:full_block}
\end{figure}

\begin{figure}[!htb]
      \centering
        \begin{tabular}{| p{5mm} | p{8mm} | p{8mm} | p{8mm} |  p{8mm} | }
   		\hline
   		\cellcolor{green!25}{1 } &  \cellcolor{blue!25}{0.2} & \cellcolor{blue!25}{0.95} & \cellcolor{blue!25} {$1e^{-4}$
		} & \cellcolor{blue!25} {0.03}\\
		\hline
 	  \end{tabular}	
	  
	  \begin{tabular}{| p{5mm} | p{8mm} | p{8mm} | p{8mm} |  p{8mm} | }
   		\hline
   		\cellcolor{green!25}{2 } & \cellcolor{blue!25}{0.1} & \cellcolor{blue!25}{0.9} & \cellcolor{blue!25} {$5e^{-4}$
		} & \cellcolor{blue!25} {0} \\
		\hline
 	  \end{tabular}	  
	  	   \caption{Example of an optimizer block. Its neighbor is obtained by selecting the next optimizer and by initializing the associated variables to their default values.}
		   \label{fig:opt_block}
\end{figure}

%-----------------------------------------------------%
\subsubsection{Neighborhood structure}
%-----------------------------------------------------%

The extended poll of \mads with categorical variables constructs one or more neighbor points from any given point, and evaluates them.
There are up to three categorical variables that are exploited using an ad~hoc generation of neighbor points. A neighborhood structure considering coupled effect between the categorical variables may find promising search spaces but it certainly increases the resources needed to carry out the optimization. To limit the number of neighbor points, the neighborhood structure considers each categorical variable independently. Hence, to create a neighbor point related to a given block, all the remaining values are fixed at the current iterate values. 
The neighbor structure of the convolution block is obtained by adding and subtracting a group of associated variables at the right of the block.  These operations can only be performed if the resulting size is within the bounds for the variable $n_1$. When adding a group of associated variables, the values of the associated variables are copied from the most right group. Adding or subtracting a group to the convolution block is illustrated in~Figure~\ref{fig:conv_block}.
The neighbor structure of the fully connected block is obtained by adding and subtracting one associated variable at the left of the block. These operations can only be performed if the resulting size is within the bounds for the variable $n_2$. When adding a group, the value of the associated variable is copied and inserted from the most left value (see example in Figure~\ref{fig:full_block}).
The structure of the network varies when adding or subtracting a convolutional or a full layer, and so does the dimension of the HPO problem. Varying the remaining categorical variables has not such effect.
The categorical variable controlling the choice of optimizer has four possible values: SGD, Adam, Adagrad or RMSprop. The choice of optimizer does not change the dimension of the optimization problem but it affects the interpretation of the four associated variables related to the optimizer as illustrated in Table~\ref{tab:optimizer}. A single neighbor point is obtained by looping between the optimizers listed in Table~\ref{tab:optimizer} from top to bottom. For each possible optimizer, there are four associated variables controlling the algorithm with different interpretations. When the optimizer is changed, these variables are reset to their initial values.
In some cases, a variable controlling a category can be well handled as an integer variable by ordering the categories with a predefined order.  This is the case for the variable selecting among the three possible activation functions (see Section~\ref{sec:hpos} and Table~\ref{tab:HPNN}): ReLU, Sigmoid and Tanh, with corresponding values between 1 and 3.   
The choice of activation function and the remaining variables are not treated as categorical variables and no neighborhood structure is required.

%------------------------------------------------%
\section{Computational results}
\label{sec-results}
%------------------------------------------------%

This section summarizes the results obtained by  \hypernomad and compares them to other methods when applied to the MNIST~\cite{mnist} and CIFAR-10~\cite{krizhevsky2009learning} data sets. For both series of tests, all the hyperparameters discussed in Section~\ref{sec:exp} are allowed to vary. However,  the user of the framework can fix some hyperparameters and choose to focus on others as described in Appendix~\ref{sec-appendix}.
All the following tests are allowed a maximum of 100 blackbox evaluations due to time limitations since one call to the blackbox takes, in average, three to four hours.

%------------------------------------------------%
\subsection{MNIST}
%------------------------------------------------%

The first tests are performed on the same blackbox provided by the authors of~\cite{Nannicini_HPO} which considers the MNIST data set with the {\sf Caffe} library~\cite{jia2014caffe}.
The \nomad software is directly used instead of \hypernomad,
 in order to compare the different methods of Table~\ref{tab:res_mnist}. The blackbox takes a simplified set of hyperparameters as described in Table~\ref{tab:simpl_hpo}, constructs a convolutional neural network that is trained on the MNIST data set~\cite{mnist},
and finally returns the validation accuracy as a measure of performance. 

\begin{table}[ht]
\caption{Hyperparameters considered for the  tests on the MNIST data set with the simplified {\sf Caffe} blackbox.}
  \centering
    \begin{tabular}{p{0.08 \textwidth} p{0.45 \textwidth} p{0.15 \textwidth} p{0.25 \textwidth}  }  \toprule
   	 {\# } & \textbf{Hyperparameter} & \textbf{Type} &  \hspace*{1cm}\textbf{Scope} \\ \midrule
	1 & Number of convolutional layers & Categorical & \hspace*{1cm}$\{$0, 1, 2$\}$\\
	2 & Number of output channels & Integer &  \hspace*{1cm}$\{$1, 2, \ldots, 50$\}$\\ 
	3 & Number of full layers & Categorical  &  \hspace*{1cm}$\{$0, 1, 2$\}$\\
	4 & Size of the full layer & Integer &  \hspace*{1cm}$\{$1, 2, \ldots, 50$\}$\\ 
	5 & Learning rate & Real &  \hspace*{1cm}[0;1] \\
	6 & Momentum    & Real 	&  \hspace*{1cm}[0;1] \\
	7 & Weight decay & Real 	&  \hspace*{1cm}[0;1] \\
	8 & Learning decay & Real  &  \hspace*{1cm}[0;1] \\\bottomrule
   \end{tabular}
   \label{tab:simpl_hpo}
\end{table}

The results are obtained by choosing five random seeds and executing the optimization five times for each seed. Table~\ref{tab:res_mnist} presents the results obtained by a random sampling (RS), RBFOpt, and SMAC, that are taken from~\cite{Nannicini_HPO}, and \nomad. These results show that using \nomad surpasses all of the other methods in terms of both the validation and the test accuracies.

%with and without the ability to explore neighboring search spaces in order to test the impact of the categorical variables and therefore the validity of the proposed approach. Concretely, disabling the categorical variables
%means that the numbers of convolutional and fully connected layers are fixed for each execution. While this is possible for a small example such as this one where there are 9 possible combinations, this approach is not recommended for bigger instances. 

\begin{table}[ht]
\caption{Results on MNIST with the simplified {\sf Caffe} blackbox.}
  \centering
   \begin{tabular}{p{0.3 \textwidth} p{0.3 \textwidth} p{0.3 \textwidth}}\toprule
   	{Algorithm } & \makecell{Average accuracy on \\  the validation set} & \makecell{Average accuracy \\ on the test set} \\ \midrule
	RS & \hspace*{2cm}94.02 & \hspace*{2cm}89.07 \\
	SMAC & \hspace*{2cm}95.48 & \hspace*{2cm}97.54 \\ 
	RBFOpt & \hspace*{2cm}95.66 & \hspace*{2cm}97.93\\
	\nomad  & \hspace*{2cm}\textbf{96.81} & \hspace*{2cm}\textbf{97.98} \\ 
	%\nomad with categorical variables & \textbf{97.54} & \textbf{97.95}\\ 
	\bottomrule
   \end{tabular}
   \label{tab:res_mnist}
\end{table}

\begin{figure}[H]
   \centering
       \includegraphics[page=1,width=.5\textwidth]{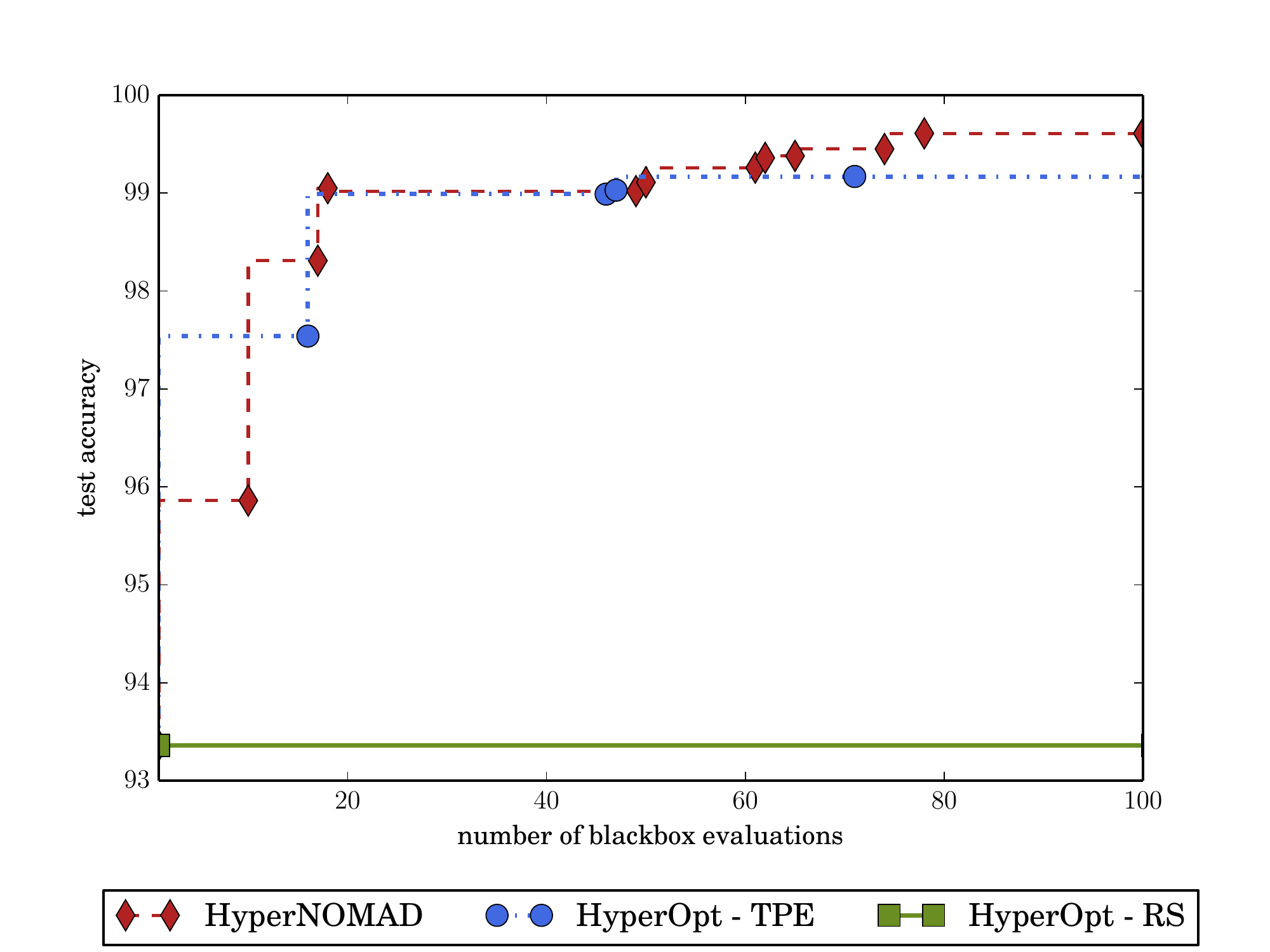} 
 \caption{Comparison between \hypernomad, TPE and RS when launched from the default starting point of \hypernomad,
 on the MNIST data set.}
 \label{fig:mnist}
\end{figure}

The next phase consists to test \hypernomad,
this time with \pytorch and its embedded MNIST data set,
and to compare it with other methods such as a random search and a Bayesian method. The \hyperopt library~\cite{bergstra2013making} is the one used in this comparison since it contains a random search in addition to TPE, a Bayesian method that relies on Parzen trees~\cite{bergstra2011algorithms}.  The blackbox used for this comparison is the one embedded in \hypernomad which allows for a greater flexibility than the previous {\sf Caffe} blackbox since it takes into account all of the hyperparameters described in Section~\ref{sec:hpos}.
The optimization is launched from the same point that corresponds to the default values for the hyperparameters in \hypernomad. This initial point contains 22 hyperparameters and obtains a test accuracy of 93.36\%. Figure~\ref{fig:mnist} shows the evolution of \hypernomad versus the two variants of \hyperopt (RS and TPE) for this test where, after 100 blackbox evaluations,  \hypernomad finds the best configuration with a final test accuracy of 99.61\%. The best solution found by TPE obtains a test accuracy of 99.17\% and the random search fails to improve the initial point.

%------------------------------------------------%
\subsection{CIFAR-10}
%------------------------------------------------%

\begin{figure}[ht]
   \centering
       \includegraphics[scale=0.35]{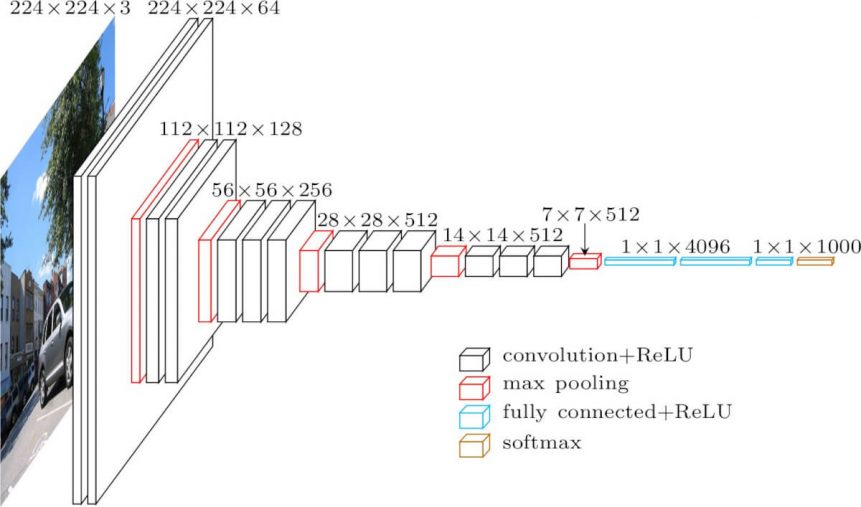}
 \caption{Architecture of the VGG-16 network. Image taken from~\cite{vgg_architecture}.}
 \label{fig:vgg16}
\end{figure}

Similarly to the previous test, \hypernomad is compared to TPE and the random search. These tests are launched using different starting points, the first being the default values of the hyperparameters in \hypernomad with 22 hyperparameters and the second being a network with the VGG-13 architecture. The VGG networks~\cite{Simonyan14c} are very deep convolutional neural networks with small kernels. Figure~\ref{fig:vgg16} illustrates the architecture of the VGG 16 network.

\begin{figure}[ht]
   \centering
      \begin{subfigure}{0.49\textwidth}
	\includegraphics[width=\linewidth]{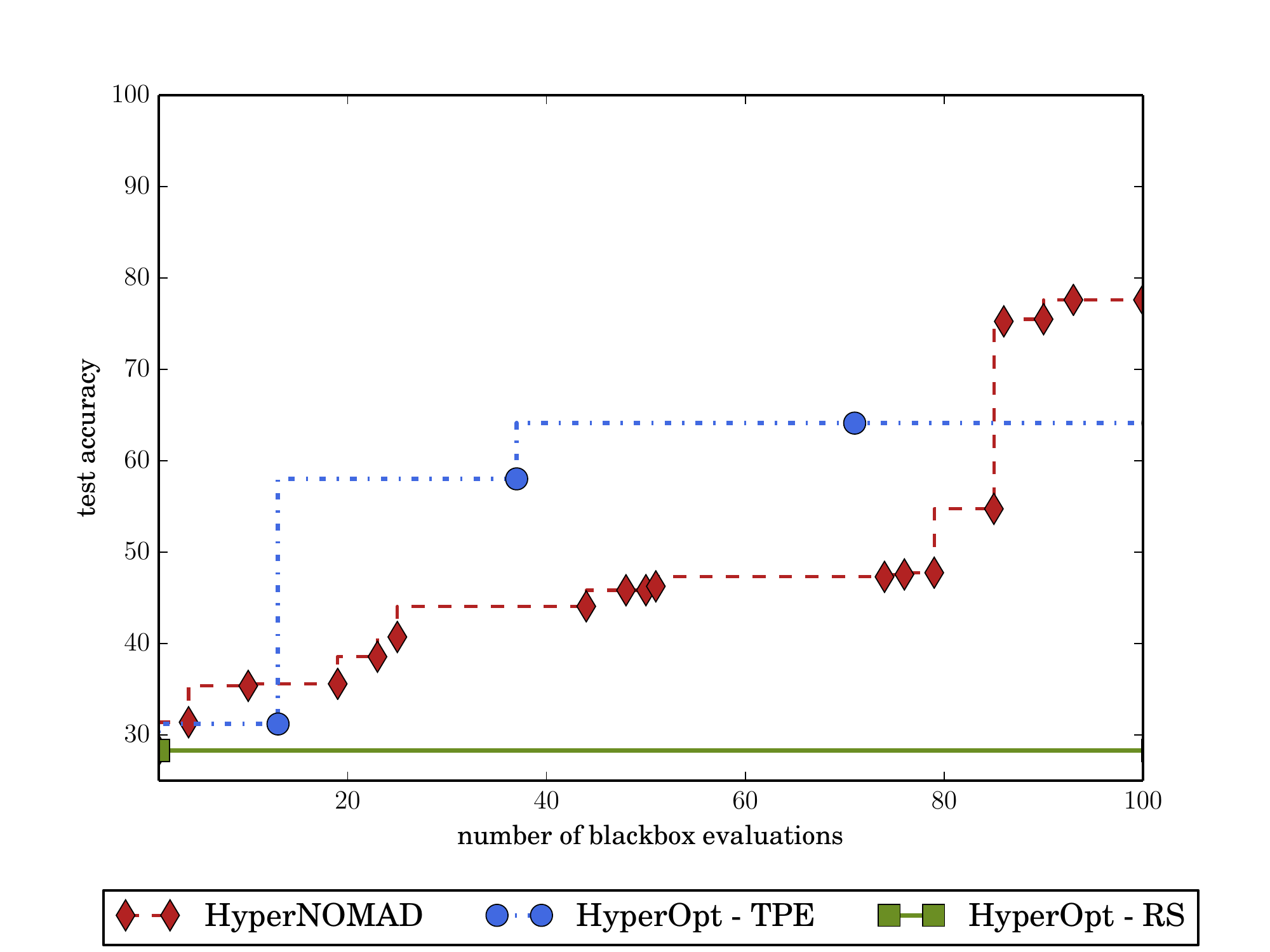}
	\caption{Default starting point.} \label{fig:comp}
   \end{subfigure}   
    \begin{subfigure}{0.49\textwidth}
	\includegraphics[width=\linewidth]{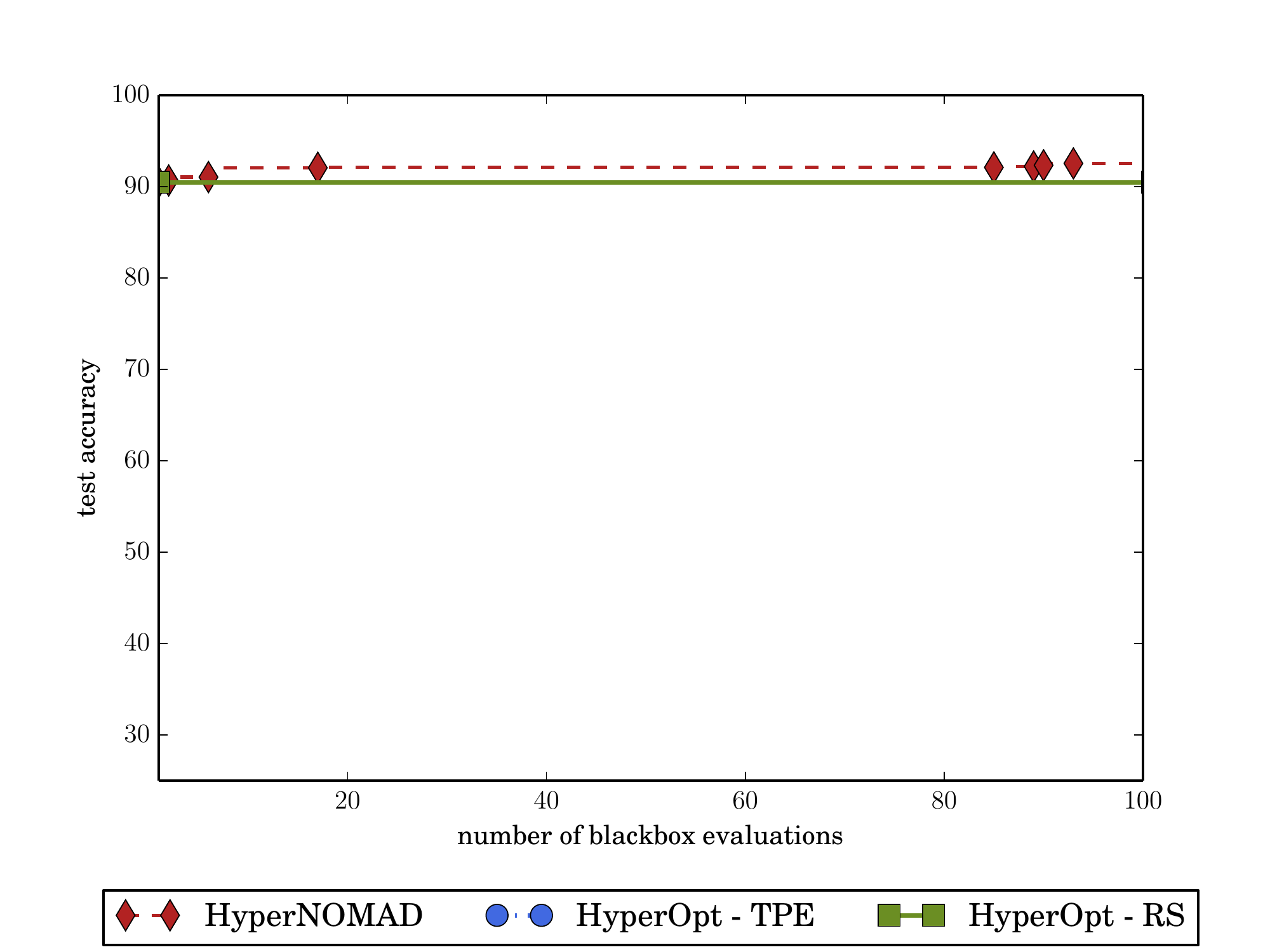}
	\caption{Starting from a VGG architecture.} \label{fig:HN_vgg13}
   \end{subfigure}
   \hspace*{\fill} % separation between the subfigures

\caption{Comparison between \hypernomad, TPE and RS, on the CIFAR-10 data set.}
 \label{fig:num_res_cifar10}
\end{figure}

Figure~\ref{fig:comp} compares \hypernomad, TPE and the random search starting from the default settings of \hypernomad which achieve a test accuracy of 28.3\%. Once again, the random search could not bring any improvements to the initial point whereas the best solution of TPE obtains a test accuracy of 64.12\% and \hypernomad finds a solution that achieves 77.6\%.

Figure~\ref{fig:HN_vgg13} shows the results of a second test performed using a starting point with a VGG-13 architecture, which corresponds to 62 hyperparameters, that achieves a test accuracy of 90.8\%. In this example neither the random search nor TPE are able to improve on the initial point given. Moreover, they spend all their evaluation budget sampling non feasible architectures. An architecture is infeasible when the size of the image passed through the convolutional layers becomes nil. This behavior can be explained by the sampling strategy of both methods since they tend to change multiple hyperparameters at once thus increasing the probability of obtaining a non feasible architecture.  \hypernomad is much more conservative when choosing a new point to evaluate which is why 50 evaluated points are feasible.
The best configuration found by \hypernomad achieves a final test accuracy of 92.54\%.

%------------------------------------------------%
\section{Discussion}
%------------------------------------------------%

This work introduces \hypernomad, a framework package for hyperparameter optimization of DNNs using the \nomad software~\cite{Le09b}. The key aspects of this framework is its ability to optimize both the architecture and the optimization phase of a deep neural network simultaneously on the one hand, and to explore different search spaces during a single execution by taking advantage of categorical variables. The framework obtains good results for both the MNIST and CIFAR-10 data sets and finds better solutions than TPE and a random search as is illustrated in Figure~\ref{fig:comp}.
Future work aims at considering different techniques of data augmentation as additional hyperparameters of the blackbox, adding more flexibility in the way the learning rate is updated and expanding the framework to other types of problems than classification and provide interfaces compatible with other tools such as {\sf Tensorflow} or {\sf Caffe2}.

%------------------------------------------------%
\section*{Acknowledgments}
%------------------------------------------------%

The authors would like to thank the Nvidia GPU Grant Program for providing a GPU used in this research and Dr. Giacomo Nannicini for providing the initial blackbox used for the preliminary testings of \hypernomad.